\def\year{2021}\relax
\documentclass[letterpaper]{article} 
\usepackage{aaai21}  
\usepackage{times}  
\usepackage{helvet} 
\usepackage{courier}  
\usepackage[hyphens]{url}  
\usepackage{graphicx} 
\usepackage{natbib}  
\usepackage{caption} 
\usepackage{amsmath}
\usepackage{amssymb}
\usepackage{algorithm}
\usepackage{algorithmic}
\usepackage{graphicx}
\usepackage{subfigure}

\newtheorem{assumption}{Assumption}
\newtheorem{definition}{Definition}
\newtheorem{theorem}{Theorem}
\newtheorem{lemma}{Lemma}

\title{Stable Adversarial Learning under Distributional Shifts}
\author{
	Jiashuo Liu\textsuperscript{\rm 1}, Zheyan Shen\textsuperscript{\rm 1}, Peng Cui\textsuperscript{\rm 1}, Linjun Zhou\textsuperscript{\rm 1}, Kun Kuang\textsuperscript{\rm 2}, Bo Li\textsuperscript{\rm 1}, Yishi Lin\textsuperscript{\rm 3}\\
}
\affiliations{
    \textsuperscript{\rm 1}Tsinghua University\\
     \textsuperscript{\rm 2}Zhejiang University\\
     \textsuperscript{\rm 3} Tencent\\

    liujiashuo77@gmail.com, shenzy17@mails.tsinghua.edu.cn, cuip@tsinghua.edu.cn, zhoulj16@mails.tsinghua.edu.cn,  kunkuang@zju.edu.cn, libo@sem.tsinghua.edu.cn, yishilin14@gmail.com

}

\begin{document}

\maketitle

\begin{abstract}
Machine learning algorithms with empirical risk minimization are vulnerable under distributional shifts due to the greedy adoption of all the correlations found in training data. 
Recently, there are robust learning methods aiming at this problem by minimizing the worst-case risk over an uncertainty set. 
However, they equally treat all covariates to form the decision sets regardless of the stability of their correlations with the target, resulting in the overwhelmingly large set and low confidence of the learner.
In this paper, we propose Stable Adversarial Learning (SAL) algorithm that leverages heterogeneous data sources to construct a more practical uncertainty set and conduct differentiated robustness optimization, where covariates are differentiated according to the stability of their correlations with the target. 
We theoretically show that our method is tractable for stochastic gradient-based optimization and provide the performance guarantees for our method. 
Empirical studies on both simulation and real datasets validate the effectiveness of our method in terms of uniformly good performance across unknown distributional shifts.
\end{abstract}

\section{Introduction}
Traditional machine learning algorithms which optimize the average loss often suffer from the poor generalization performance under distributional shifts induced by latent heterogeneity, unobserved confounders or selection biases in training data\cite{daume2006domain,torralba2011unbiased, kuang2018stable,shen2019stable}. 
However, in high-stake applications such as medical diagnosis\cite{kukar2003transductive}, criminal justice\cite{berk2018fairness,rudin2018optimized} and autonomous driving \cite{huval2015empirical}, it is critical for the learning algorithms to ensure the robustness against potential unseen data.
Therefore, robust learning methods have recently aroused much attention due to its favorable property of robustness guarantee\cite{ben1998robust,goodfellow2014explaining,madry2017towards}.

Instead of optimizing the empirical cost on training data, robust learning methods seek to optimize the worst-case cost over an uncertainty set and can be further separated into two main branches named adversarially and distributionally robust learning. 
In adversarially robust learning, the uncertainty set is constructed point-wisely\cite{goodfellow2014explaining,papernot2016limitations,madry2017towards,ye2018bayesian}. 
Adversarial attack is performed independently on each data point within a $L_2$ or $L_{\infty}$ norm ball around itself. 
In distributionally robust learning, on the other hand, the uncertainty set is characterized on a distributional level\cite{SinhaCertifying, esfahani2018data,duchi2018learning}. 
A joint perturbation, typically measured by Wasserstein distance or $f$-divergence, is applied to the entire distribution entailed by training data.
These methods can provide robustness guarantees under distributional shifts when testing distribution is captured in the uncertainty set. 
However, in real scenarios, to contain the true distribution, the uncertainty set is often overwhelmingly large, which is also referred to as the over pessimism or the low confidence problem\cite{frogner2019incorporating,sagawa2019distributionally}.  
Specifically, with an overwhelmingly large set, the learner optimizes for implausible worst-case scenarios, resulting in meaningless results (e.g. the classifier assigns equal probability to all classes).
Such a problem greatly hurts the generalization ability of robust learning methods in practice.

The essential problem of the above methods lies in the construction of the uncertainty set.
To address the over pessimism of the learning algorithm, one should form a more practical uncertainty set which is likely to contain the potential distributional shifts in the future.
More specifically, in real applications we observe that different covariates may be perturbed in a non-uniform way, which should be considered in building a practical uncertainty set. 
Taking the problem of waterbirds and landbirds classification as an example\cite{wah2011caltech}.
There exist two types of covariates where the stable covariates (e.g. representing the bird itself) preserve immutable correlations with the target across different environments, while those unstable ones (e.g. representing the background) are likely to change.
Therefore, for the example above, the construction of the uncertainty set should mainly focus on the perturbation of those unstable covariates (e.g. background) to generate more practical and meaningful samples. 
Following such intuition, there are several work\cite{DBLP:journals/corr/abs-1904-06347,vaishnavi2019can} based on the adversarial attack which focus on perturbing the color or background of images to improve the adversarial robustness. 
However, these methods mainly follow a step by step routine where the segmentation is conducted first to separate the background from the foreground and cannot theoretically provide robustness guarantees under unknown distributional shifts, which limits their applications on more general settings.

In this paper, we propose the Stable Adversarial Learning (SAL) algorithm to address this problem in a more principled and unified way, which leverages heterogeneous data sources to construct a more practical uncertainty set.
Specifically, we adopt the framework of Wasserstein distributionally robust learning(WDRL) and further characterize the uncertainty set to be anisotropic according to the stability of covariates across the multiple environments, which induces stronger adversarial perturbations on unstable covariates than those stable ones.  
A synergistic algorithm is designed to jointly optimize the covariates differentiating process as well as the adversarial training process of model's parameters. 
Compared with traditional robust learning techniques, the proposed method is able to provide robustness under strong distributional shifts while maintaining enough confidence of the learner. 
Theoretically, we prove that our method constructs a more compact uncertainty set, which as far as we know is the first analysis of the compactness of adversarial sets in WDRL literature.
Empirically, the advantages of our SAL algorithm are demonstrated on both synthetic and real-world datasets in terms of uniformly good performance across distributional shifts.

\section{The SAL Method}
We first introduce the Wasserstein Distributionally Robust Learning (WDRL) framework which attempts to learn a model with minimal risk against the worst-case distribution in the uncertainty set characterized by Wasserstein distance: 
\begin{definition}
Let $\mathcal{Z} \subset \mathbb{R}^{m+1}$ and $\mathcal{Z} = \mathcal{X}\times\mathcal{Y}$ , given a transportation cost function $c: \mathcal{Z} \times \mathcal{Z} \rightarrow [0, \infty)$, which is nonnegative, lower semi-continuous and satisfies $c(z,z)=0$, for probability measures $P$ and $Q$ supported on $\mathcal{Z}$, the Wasserstein distance between $P$ and $Q$ is :
\begin{equation}
W_c(P, Q) = \inf\limits_{M \in \Pi(P,Q)} \mathbb{E}_{(z,z') \sim M}[c(z,z')]
\end{equation}
where $\Pi(P,Q)$ denotes the couplings with $M(A,\mathcal{Z})=P(A)$ and $M(\mathcal{Z},A)=Q(A)$ for measures $M$ on $\mathcal{Z}\times \mathcal{Z}$.
\end{definition}

As mentioned above, the uncertainty set built in WDRL is often overwhelmingly large in wild high-dimensional scenarios. 
To demonstrate this over pessimism problem of WDRL, we design a toy example in \ref{exp:toy} to show the necessity to construct a more practical uncertainty set. 
Indeed, without any prior knowledge or structural assumptions, it is quite difficult to design a practical set for robustness under distributional shifts.
Therefore, we consider a more flexible setting with heterogeneous datasets $D^e = \{X^e,Y^e\}$ from multiple training environments $e \in \mathcal{E}_{tr}$.
Specifically, each dataset $D^e$ contains examples identically and independently distributed according to some joint distribution $P_{XY}^e$ on $\mathcal{X}\times\mathcal{Y}$.
Then we come up with one basic assumption for our problem.
Given the observations that in real scenarios, different covariates have different extents of stability, we propose assumption \ref{assup1}. 
\begin{assumption}
\label{assup1}
	There exists a decomposition of all the covariates $X = \{S,V\}$, where $S$ represents the stable covariate set and V represents the unstable one, so that for all environments $e \in \mathcal{E}$, $\mathbb{E}[Y^e|S^e=s,V^e=v] = \mathbb{E}[Y^e|S^e=s] = \mathbb{E}[Y|S=s]$. 
\end{assumption}

Intuitively, assumption \ref{assup1} indicates that the correlation between stable covariates $S$ and the target $Y$ stays invariant across environments, which is quite similar to the assumption in \cite{arjovsky2019invariant,kuang2020stable, shen2020stable}. 
Moreover, assumption \ref{assup1} also demonstrates that the influence of $V$ on the target $Y$ can be wiped out as long as whole information of $S$ is accessible.
Under the assumption \ref{assup1}, the disparity among covariates revealed in the heterogeneous datasets can be leveraged for better construction of the uncertainty set. 

Here we propose the Stable Adversarial Learning (SAL) algorithm, which leverages heterogeneous data to build a more practical uncertainty set with covariates differentiated according to their stability. 
The objective function of our SAL algorithm is:
\begin{small}
\begin{align}
\label{equ:obj1}
	&\min\limits_{\theta\in\Theta}\sup_{Q: W_{c_w}(Q,P_0)\leq \rho}\mathbb{E}_{X,Y\sim Q}[\ell(\theta;X,Y)]\\
\label{equ:cw}
	\text{where}\ c_w&(z_1, z_2) = \|w\odot(z_1-z_2)\|_2^2\ \ \text{and}   \\
\label{equ:obj2}
	w \in \arg&\min\limits_{w\in \mathcal{W}} \left\{ \frac{1}{|\mathcal{E}_{tr}|}\sum_{e\in\mathcal{E}_{tr}}\mathcal{L}^e(\theta) + \alpha\max\limits_{e_p,e_q\in \mathcal{E}_{tr}}\mathcal{L}^{e_p} - \mathcal{L}^{e_q}\right\}
\end{align}
\end{small}
where $P_0$ denotes the training distribution, $W_{c_w}$ denotes the Wasserstein distance with transportation cost function $c_w$ defined as equation \ref{equ:cw}, $\mathcal{W} = \left\{w : w\in[1,+\infty)^{m+1} \ \ \&\&\  \min(w^{(1)}, \dots, w^{(m+1)})=1\right\}$ denotes the covariate weight space($w^{(i)}$ denotes the $i$th element of $w$), 
and $\mathcal{L}^e$ denotes the average loss in environment $e\in\mathcal{E}_{tr}$, $\alpha$ is a hyper-parameter to adjust the tradeoff between average performance and the stability.

Intuitively, $w$ controls the perturbation level of each covariate and formulates an anisotropic uncertainty set compared with the conventional WDRL methods.
The objective function of $w$ (equation \ref{equ:obj2}) contains two parts: the average loss in training environments as well as the maximum margin, which aims at learning such $w$ that the resulting uncertainty set   leads to a learner with uniformly good performance across environments. 
Equation \ref{equ:obj1} is the objective function of model's parameters via distributionally robust learning with the learnable covariate weight $w$. 
During training, the covariate weight $w$ and model's parameters $\theta$ are iteratively optimized.
Details of the algorithm are delineated below. We first will introduce the optimization of model's parameter in section \ref{sec4-1:optimization}, then the transportation cost function learning procedure in section \ref{sec4-2:weight}. 

\subsection{Tractable Optimization}
\label{sec4-1:optimization}
In SAL algorithm, the model's parameters $\theta$ and covariate weight $w$ is optimized iteratively. In each iteration, given current $w$, the objective function for $\theta$ is:
\begin{equation}
\label{equ:sec4-obj1}
	\min\limits_{\theta\in\Theta}\sup_{Q: W_{c_w}(Q,P_0)\leq \rho}\mathbb{E}_{X,Y\sim Q}[\ell(\theta;X,Y)]
\end{equation}
The duality results in lemma \ref{theorem_refor} show that the infinite-dimensional optimization problem (\ref{equ:sec4-obj1}) can be reformulated as a finite-dimensional convex optimization problem \cite{esfahani2018data}. Besides, inspired by \cite{SinhaCertifying}, a Lagrangian relaxation is provided for computation efficiency.

\begin{lemma}
\label{theorem_refor}
Let $\mathcal{Z} = \mathcal{X}\times\mathcal{Y}$ and $\ell: \Theta \times \mathcal{Z} \rightarrow \mathbb{R}$ be continuous. For any distribution $Q$ and any $\rho \ge 0$, let $s_{\lambda}(\theta;(x,y)) = \sup\limits_{\xi \in \mathcal{Z}} (\ell(\theta;\xi) - \lambda c_w(\xi,(x,y)))$, $\mathcal{P}=\{Q: W_c(Q,P_0)\leq \rho\}, $we have:
	\begin{equation}
	\label{duality}
		\sup\limits_{Q \in \mathcal{P}} \mathbb{E}_Q[\ell(\theta;x,y)] = \inf\limits_{\lambda \geq 0} \{\lambda\rho + \mathbb{E}_{P_0}[s_{\lambda}]\}
	\end{equation}
and for any $\lambda \geq 0$, we have:
\begin{equation}
		\label{equ:relax}
		\sup\limits_{Q \in \mathcal{P}}\{\mathbb{E}_Q[\ell(\theta;(x,y))]-\lambda W_{c_w}(Q,P_0)\} =\mathbb{E}_{P_0}[s_{\lambda}] 
	\end{equation}
\end{lemma}

Notice that there exists only the inner supremum in $\mathbb{E}_{P_0}[s_{\lambda}(\theta;(x,y))]$, which can be seen as a relaxed Lagrangian penalty function of the original objective function (\ref{equ:sec4-obj1}). 
Here we give up the prescribed amount $\rho$ of robustness in equation (\ref{equ:sec4-obj1}) and focus instead on the relaxed Lagrangian penalty function for efficiency in equation (\ref{equ:relax}). 
The loss function on empirical distribution $\hat{P_N}$ becomes $\frac{1}{N}\sum_{i=1}^N s_{\lambda}(\theta;(x_i,y_i))$.
We adopt adversarial training procedure proposed in \cite{SinhaCertifying} to approximate the supremum for $s_{\lambda}$. 

Specifically, given predictor $x$, we adopt gradient ascent to obtain an approximate maximizer $\hat{x}$ of $\left\{\ell(\theta;(\hat{x},y))-\lambda c_w(\hat{x},x)\right\}$ and optimize the model's parameter $\theta$ using $\hat{x}$ as: $\hat{\mathcal{L}} = \frac{1}{N}\sum_{i=1}^N \ell(\theta;\hat{x},y)$.
In the following parts, we simply use $X_A$ to denote $\{\hat{x}\}_N$, which means the set of maximizers for training data $\{x\}_N$. 
The convergence guarantee for this optimization can be referred to \cite{SinhaCertifying}.

\subsection{Learning for Transportation Cost Function}
\label{sec4-2:weight}

We introduce the learning for transportation cost function $c_w$ in this section. In supervised scenarios, perturbations are typically only added to predictor $X$ and not target $Y$. Therefore, we simplify $c_w: \mathcal{Z}\times\mathcal{Z}\rightarrow[0,+\infty)(\mathcal{Z}=\mathcal{X}\times\mathcal{Y})$ to be:
\begin{small}
\begin{align}
\label{equ:w-simplify}
	c_w(z_1,z_2) &= c_w(x_1,x_2) + \infty\times\mathbb{I}(y_1\neq y_2)\\
				 &= \|w\odot(x_1-x_2)\|_2^2 + \infty\times\mathbb{I}(y_1\neq y_2)
\end{align}
\end{small}
and omit '$y$-part' in $c_w$ as well as $w$, that is $w \in [1,+\infty)^{m}$  in the following parts. 
Intuitively, $w$ controls the strength of adversary put on each covariate. 
The higher the weight is, the weaker perturbation is put on the corresponding covariate. 
Ideally, we hope the covariate weights on stable covariates are extremely high to protect them from being perturbed and to maintain the stable correlations, while weights on unstable covariates are nearly $1$ to encourage perturbations for breaking the harmful spurious correlations. 
With the goal towards uniformly good performance across environments, we come up with the objective function $R(\theta(w))$ for learning $w$ as:
\begin{small}
\begin{equation}
	R(\theta(w))=\frac{1}{|\mathcal{E}_{tr}|}\sum_{e\in\mathcal{E}_{tr}}\mathcal{L}^e(\theta(w)) + \alpha\max\limits_{e_p,e_q\in \mathcal{E}_{tr}}\left(\mathcal{L}^{e_p} - \mathcal{L}^{e_q}\right)
\end{equation}
\end{small}
where $\alpha$ is the hyper-parameter. $R(\theta(w))$ contains two parts: the first is the average loss in multiple training environments; the second reflects the max margin among environments, which reflects the stability of $\theta(w)$, since it is easy to prove that $\max\limits_{e_p,e_q\in \mathcal{E}_{tr}}\mathcal{L}^{e_p}(\theta(w)) - \mathcal{L}^{e_q}(\theta(w))=0$ if and only if the errors among all training environments are same. 
Here $\alpha$ is used to adjust the tradeoff between average performance and stability. 

In order to optimize $w$, $\partial R(\theta(w))/\partial w$ can be approximated as following. 
\begin{equation}
	\small
	\frac{\partial R(\theta(w))}{\partial w} = \frac{\partial R}{\partial \theta}\frac{\partial \theta}{\partial X_A}\frac{\partial X_A}{\partial w}
\end{equation}
Note that the first term $\partial R/ \partial \theta$ can be calculated easily. The second term can be approximated during the gradient descent process of $\theta$ as :
\begin{equation}
\small
	\frac{\partial \theta}{\partial X_A} \approx -\epsilon\sum_t \frac{\nabla_{\theta}\hat{\mathcal{L}}(\theta^t;X_A,Y)}{\partial X_A}
\end{equation}
where $\frac{\nabla_{\theta}\hat{\mathcal{L}}(\theta^t;X_A,Y)}{\partial X_A}$ can be calculated during the training process. The third term $\partial X_A / \partial w$ can be approximated during the adversarial learning process of $X_A$ as:
\begin{equation}
\small
	\frac{\partial X_A}{\partial w} \approx -2\epsilon_x\lambda \sum_t \mathrm{Diag}(X_A^t - X)
\end{equation}
which can be accumulated during the adversarial training process. 
Then given current $\theta$, we can update $w$ as:
\begin{equation}
\small
	w^{t+1} = Proj_{\mathcal{W}}\left(w^t - \epsilon_w \frac{\partial R(\theta^t)}{\partial w}\right)
\end{equation}
where $Proj_{\mathcal{W}}$ means projecting onto the space $\mathcal{W}$.

\section{Theoretical Analysis}
Here we first provide the robustness guarantee for our method, and then we analyze the rationality of our uncertainty set, which also demonstrates the uncertainty set built in our SAL is more practical. 
First, we provide the robustness guarantee in theorem \ref{theorem:bound} with the help of lemma \ref{theorem_refor} and Rademacher complexity\cite{bartlett2002rademacher}. 
\begin{theorem}
\label{theorem:bound}
Let $\Theta=R^m,\ x\in \mathcal{X},\ y\in\mathcal{Y}$.
Assume $|\ell(\theta;z)|$ is bounded by $T_{\ell} \geq 0$ for all $\theta\in \Theta,\ z=(x,y)\in\mathcal{X}\times\mathcal{Y}$.
Let $F:\mathcal{X}\rightarrow\mathcal{Y}$ be a class of prediction functions, then for $\theta\in \Theta,\ \rho \geq 0,\ \lambda \geq 0$, with probability at least $1-\delta$, for $P \in \{P:W_{c_w}(P,P_0)\leq \rho\}$, we have:
\begin{small}
\begin{equation}
	\sup\limits_{P}\mathbb{E}_P\left[\ell(\theta;Z)\right] \leq \lambda\rho + \mathbb{E}_{\hat{P}_n}\left[s_{\lambda}(\theta;Z)\right] + \mathcal{R}_n(\widetilde{\ell}\circ F) + kT_{\ell}\sqrt{\frac{\ln(1/\delta)}{n}}
\end{equation}
\end{small}
Specially, let $M(\theta;z_0)= \arg\min\limits_{z\in\mathcal{Z}}\left\{s_{\lambda}(\theta;z_0)\right\}$ when $\hat{\rho}_n(\theta) = \mathbb{E}_{\hat{P}_n}\left[c_w(M(\theta;Z),Z)\right]$, for $P \in \{P:W_{c_w}(P,P_0)\leq \hat{\rho}_n(\theta)\}$, 
\begin{equation}
\small
	\sup\limits_{P}\mathbb{E}_P\left[\ell(\theta;Z)\right] = \sup\limits_{P} \mathbb{E}_P\left[\ell(\theta;Z)\right] + \mathcal{R}_n(\widetilde{\ell}\circ F) + kT_{\ell}\sqrt{\frac{\ln(1/\delta)}{n}}
\end{equation}
with probability at least $1-\delta$, where $\widetilde{\ell}\circ F=\{(x,y)\mapsto \ell(f(x),y)-\ell(0,y):f\in F\}$ and $\mathcal{R}_n$ denotes the Rademacher complexity\cite{bartlett2002rademacher} and $k$ is a numerical constant no less than 0. 
\end{theorem}
Theorem \ref{theorem:bound} is the standard result on Rademacher complexity as in previous distributionally robust optimization literature. 
It proves our empirical loss given by our optimization method can control the original worst-case cost of the uncertainty set in SAL.

Then we analyze the rationality of our method in theorem \ref{theorem:sparse}, where our major theoretical contribution lies on. 
As far as we know, it is the first analysis of the compactness of adversary sets in WDRL literature.

\begin{assumption}
\label{assump:theoretical}
	Given $\rho>0$, $\exists Q_0 \in \mathcal{P}_0$ that satisfies: 
	
	$\mathrm{(1)}$\ $\forall \epsilon>0$, $\left|\inf\limits_{M \in \Pi(P_0,Q_0)}\mathbb{E}_{(z_1,z_2 \sim M)}\left[c(z_1,z_2)\right]\right| \leq \epsilon$, we refer to the couple minimizing the expectation as $M_0$.
	
	$\mathrm{(2)}$\ $\mathbb{E}_{M\in \Pi(P_0,Q_0)-M_0}\left[c(z_1,z_2)\right]\geq \rho$, where $\Pi(P_0,Q_0)-M_0$ means excluding $M_0$ from $\Pi(P_0,Q_0)$.
	
	$\mathrm{(3)}$\ $Q_{0\#S}\neq P_{0\#S}$, where $S=\{i: w^{(i)}>1\}$ and $w^{(i)}$ denotes the $i$th element of $w$ and $P_{\#S}$ denotes the marginal distribution on dimensions $S$.
\end{assumption} 

Assumption \ref{assump:theoretical} describes the boundary property of the original uncertainty set $\mathcal{P}_0 = \{Q:W_c(Q,P_o)\leq \rho\}$, which assumes that there exists at least one distribution on the boundary whose marginal distribution on $S$ is not the same as the center distribution $P_0$'s and is easily satisfied. 
Based on this assumption, we come up with the following theorem.

\begin{theorem}
\label{theorem:sparse}
Under assumption \ref{assump:theoretical}, assume the transportation cost function in Wasserstein distance takes form of $c(x_1,x_2)=\|x_1-x_2\|_1$ or $c(x_1,x_2)=\|x_1-x_2\|_2^2$. Then, given observed distribution $P_0$ supported on $\mathcal{Z}$ and $\rho \geq 0$, for the adversary set $\mathcal{P} = \{Q:W_{c_w}(Q,P_0)\leq \rho\}$ and the original $\mathcal{P}_0 = \{Q:W_{c}(Q,P_0)\leq \rho\}$, given $c_w$ where $\min(w^{(1)},\dots, w^{(m)})=1$ and $\max(w^{(1)},\dots, w^{(m)})>1$, we have $\mathcal{P} \subset \mathcal{P}_0$. Furthermore, for the set $U=\{i | w^{(i)}=1\}$, $\exists Q_0 \in \mathcal{P}$ that satisfies $W_{c_w}(P_{0\#U},Q_{0\#U})=\rho$.
\end{theorem}
Theorem \ref{theorem:sparse} proves that the constructed uncertainty set of our method is smaller than the original. 
Intuitively, in adversarial learning paradigm, if stable covariates are perturbed, the target should also change correspondingly to maintain the underlying relationship.
However, we have no access to the target value corresponding to the perturbed stable covariates in practice, so optimizing under an isotropic uncertainty set (e.g. $P_0$) which contains perturbations on both stable and unstable covariates would generally lower the confidence of the learner and produce meaningless results. 
Therefore, from this point of view, by adding high weights on stable covariates in the cost function, we may construct a more reasonable and practical uncertainty set in which the ineffective perturbations are avoided.

\section{Experiments}
In this section, we validate the effectiveness of our method on simulation data and real-world data.

{\bf Baselines\ \ \ } We compare our proposed SAL with the following methods. 
\begin{itemize}
\small
	\item Empirical Risk Minimization(ERM):\ \ \ \ \ \  $\min\limits_{\theta}\mathbb{E}_{P_0}\left[\ell(\theta;X,Y)\right]$
	\item Wasserstein Distributionally Robust Learning(WDRL):\ $\min\limits_{\theta}\sup\limits_{Q \in W(Q,P_0)\leq \rho}\mathbb{E}_{Q}\left[\ell(\theta;X,Y)\right]$
	\item Invariant Risk Minimization(IRM\cite{arjovsky2019invariant}):\ \ \ \ \ $\min\limits_{\theta}\sum_{e\in\mathcal{E}}\mathcal{L}^e+\lambda\|\nabla_{w|w=1.0}\mathcal{L}^e(w\cdot\theta)\|^2$
\end{itemize}
For ERM and WDRL, we simply pool the multiple environments data for training. 
For fairness, we search the hyper-parameter $\lambda$ in $\{0.01, 0.1, \dots, 1e0, 1e1,\dots,1e4\}$ for IRM and the hyper-parameter $\rho$ in $\{1, 5, 10, 20,50,80,100\}$ for WDRL, and select the best hyper-parameter according to the validation performance. 

{\bf Evaluation Metrics\ } To evaluate the prediction performance, we use $\mathrm{Mean\_Error}$ defined as $\mathrm{Mean\_Error} = \frac{1}{|\mathcal{E}_{te}|}\sum_{e \in \mathcal{E}_{te}}\mathcal{L}^e$ and $\mathrm{Std\_Error}$ defined as $\mathrm{Std\_Error} = \sqrt{\frac{1}{|\mathcal{E}_{te}|-1}\sum_{e\in \mathcal{E}_{te}}\left(\mathcal{L}^e - \mathrm{Mean\_Error}\right)^2}$, which are the mean and standard deviation error across testing environments $e\in\mathcal{E}_{te}$. 

{\bf Imbalanced Mixture} In our experiments, we perform a non-uniform sampling among different environments in training set which follows the natural phenomena that empirical data follow a power-law distribution. It is widely accepted that only a few environments/subgroups are common and the rest majority are rare\cite{2018Causally,sagawa2019distributionally,2020An}.

\subsection{Simulation Data}
Firstly, we design one toy example to demonstrate the over pessimism problem of conventional WDRL. 
Then, we design two mechanisms to simulate the varying correlations of unstable covariates across environments, named by selection bias and anti-causal effect. 

\subsubsection{Toy Example}
\label{exp:toy}

\begin{figure*}[ht]
\vskip -0.1in
\subfigure[Testing performance for each environment.]{
\label{img:toy-radius2}
\includegraphics[width=0.32\linewidth]{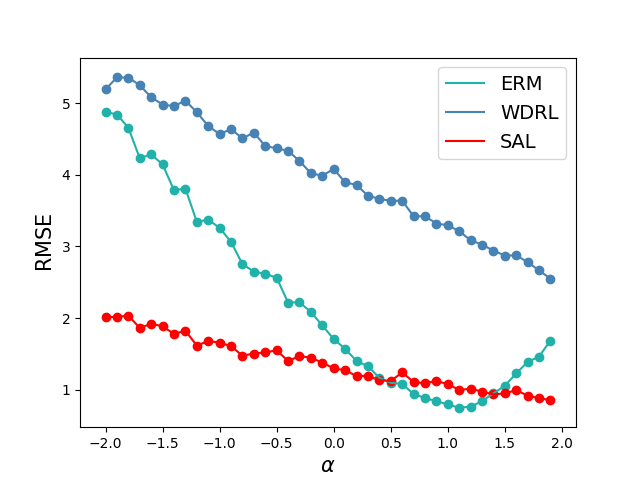}}
\subfigure[Testing performance with respect to radius]{
\label{img:toy-radius}
\includegraphics[width=0.32\linewidth]{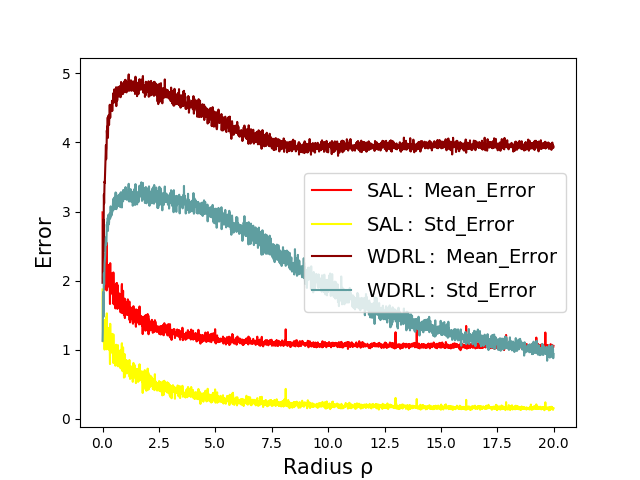}}
\subfigure[The learned coefficient value of $S$ and $V$ with respect to radius]{
\label{img:toy-estimate}
\includegraphics[width=0.32\linewidth]{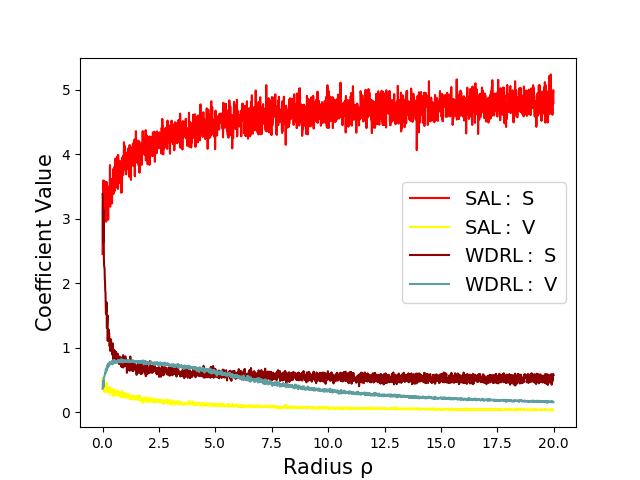}}

\caption{Results of the toy example. The left figure shows the testing performance in different environments under fixed radius, where $\mathrm{RMSE}$ is root mean square error for the prediction. The middle and right denotes the prediction error and the learned coefficients of WDRL and SAL with respect to radius respectively.}

\vskip -0.1in
\end{figure*}
\label{sec:toy}
In this setting, we have $Y= 5*S +S^2+\epsilon,\ V=\alpha Y+\epsilon$, where the effect of $S$ on $Y$ stays invariant, but the correlation between  $V$ and $Y$, i.e. the parameter $\alpha$, varies across environments.
In training, we generate 180 data points with $\alpha=1$ for environment 1 and 20 data points with $\alpha=-0.1$ for environment 2. 
We compared methods for linear regression across testing environments with $\alpha \in \{-2.0,-1.5,\dots,1.5,2.0\}$. 

We first set the radius for WDRL and SAL to be 20.0, and the results are shown in Figure \ref{img:toy-radius2}. 
We find the ERM induces high estimation error as it puts high regression coefficient on $V$. Therefore, it performs poor in terms of prediction error when there are distribution shifts. 
While WDRL achieves more robust performance than ERM across environments, the prediction error is much higher than the others.
Our method SAL achieves not only the smallest prediction error, but also the most robust performance across environments.

Furthermore, we train SAL and WDRL for linear regression with a varying radius $\rho \in \{0.0, 0.01,\dots,20.0\}$. 
From the results shown in Figure \ref{img:toy-radius}, we can see that, with the radius growing larger, the robustness of WDRL becomes better, but meanwhile, its performance maintains poor in terms of high $Mean\_Error$ and much worse than ERM ($\rho=0$). 
This further verifies the limitation of WDRL with respect to the overwhelmingly-large adversary distribution set. 
In contrast, SAL achieves not only better prediction performance but also better robustness across environments.  
The plausible reason for the performance difference between WDRL and SAL can be explained by Figure \ref{img:toy-estimate}.
As the radius $\rho$ grows larger, WDRL tends to conservatively estimate small coefficients for both $S$ and $V$ so that the model can produce robust prediction performances over the overwhelmingly-large uncertainty set. 
Comparatively, as our SAL provides a mechanism to differentiate covariates and focus on the robustness optimization over unstable ones, the learned coefficient of unstable covariate $V$ is gradually decreased to improve robustness, while the coefficient of stable covariate $S$ does not change much to guarantee high prediction accuracy.


\subsubsection{Selection Bias}
In this setting, the correlations between unstable covariates and the target are perturbed through selection bias mechanism. 
According to assumption \ref{assup1}, we assume $X=[S,V]^T$ and $Y = f(S) + \epsilon$ and $P(Y|S)$ remains invariant across environments while $P(Y|V)$ can arbitrarily change. 
For simplicity, we select data points according to a certain unstable covariate $v_0$.
\begin{align}
\hat{P}(x) = |r|^{-5*|f(s) - sign(r)*v_0|}
\end{align}  
where $|r| > 1$ and $\hat{P}(x)$ denotes the probability of point $x$ to be selected. 
Intuitively, $r$ eventually controls the strengths and direction of the spurious correlation between $v_0$ and $Y$(i.e. if $r>0$, a data point whose $v_0$ is close to its $y$ is more probably to be selected.).
The larger value of $|r|$ means the stronger spurious correlation between $v_0$ and $Y$, and $r \ge 0$ means positive correlation and vice versa. 
Therefore, here we use $r$ to define different environments.

In training, we generate $n$ data points, where $\kappa n$ points from environment $e_1$ with a predefined $r$ and $(1-\kappa)n$ points from $e_2$ with $r=-1.1$. In testing, we generate data points for 10 environments with $r \in [-3,-2,-1.7,\dots,1.7,2,3]$. $\beta$ is set to 1.0. 

\begin{table*}[htbp]
	\small
	\centering
	
	\vskip 0.05in
	
	\begin{tabular}{|l|c|c|c|c|c|c|}
		\hline
		\multicolumn{7}{|c|}{\textbf{Scenario 1: varying selection bias rate $r$\quad($n=2000,p=10,\kappa=0.95$)}}\\
		\hline
		$r$&\multicolumn{2}{|c|}{$r=1.5$}&\multicolumn{2}{|c|}{$r=1.7$}&\multicolumn{2}{|c|}{$r=2.0$}\\
		\hline
		Methods &  $\mathrm{Mean\_Error}$ & $\mathrm{Std\_Error}$ &$\mathrm{Mean\_Error}$ & $\mathrm{Std\_Error}$ &  $\mathrm{Mean\_Error}$ & $\mathrm{Std\_Error}$  \\
		\hline 
		ERM & 0.484 & 0.058 & 0.561 & 0.124  & 0.572 & 0.140  \\
		WDRL& 0.482 & 0.044& 0.550 & 0.114  & 0.532 & 0.112 \\
		IRM & 0.475 &\bf 0.014 & 0.464 &\bf 0.015 & 0.477 &\bf 0.015 \\
		\hline
		SAL &\bf 0.450 & 0.019 &\bf  0.449 &\bf  0.015  &\bf 0.452 & 0.017 \\
		\hline
		\multicolumn{7}{|c|}{\textbf{Scenario 2: varying ratio $\kappa$ and sample size $n$\quad($p=10,r = 1.7$)}}\\
		\hline
		$\kappa,n$&\multicolumn{2}{|c|}{$\kappa=0.90, n=500$}&\multicolumn{2}{|c|}{$\kappa=0.90, n=1000$}&\multicolumn{2}{|c|}{$\kappa=0.975, n=4000$}\\
		\hline
		Methods &$\mathrm{Mean\_Error}$ & $\mathrm{Std\_Error}$ &$\mathrm{Mean\_Error}$ & $\mathrm{Std\_Error}$ &   $\mathrm{Mean\_Error}$ & $\mathrm{Std\_Error}$ \\
		\hline %
		ERM & 0.580 & 0.103 & 0.562 & 0.113  & 0.555 & 0.110 \\ 
		WDRL & 0.563 & 0.101& 0.527 & 0.083 & 0.536 & 0.108 \\
		IRM & 0.460 &\bf 0.014 & 0.464 &\bf 0.015 & 0.459	&\bf 0.014\\
		\hline
		SAL &\bf 0.454 & 0.015 & \bf 0.451 & \bf 0.015 & \bf 0.448 & \bf 0.014\\
		\hline

	\end{tabular}

\caption{Results in selection bias simulation experiments of different methods with varying selection bias $r$, ratio $\kappa$ and sample size $n$ of training data, and each result is averaged over ten times runs.}
\label{tab:simulation2}
\end{table*}

We compare our SAL with ERM, IRM and WDRL for Linear Regression. We conduct extensive experiments with different settings on $r$, $n$, and $\kappa$. 
In each setting, we carry out the procedure 15 times and report the average results. 
The results are shown in Table \ref{tab:simulation2}.  

From the results, we have the following observations and analysis:
{\bf ERM\ } suffers from the distributional shifts in testing and yields poor performance in most of the settings.
Compared with ERM, the other three robust learning methods achieve better average performance due to the consideration of robustness during the training process.
When the distributional shift becomes serious as $r$ grows, {\bf WDRL} suffers from the overwhelmingly-large distribution set and performs poorly in terms of prediction error, which is consistent with our analysis.
{\bf IRM} has stable performances across testing environments, while its average error is higher than SAL, which reveals that IRM may harm the average performance for stability. 
Compared with other robust learning baselines, our {\bf SAL} achieves nearly perfect performance with respect to average performance and stability, especially the variance of losses across environments close to 0, which reflects the effectiveness of assigning different weights to covariates for constructing the uncertainty set.

\subsubsection{Anti-causal Effect}
Inspired by \cite{arjovsky2019invariant}, in this setting, we introduce the spurious correlation by using anti-causal relationship from the target $Y$ to the unstable covariates $V$.
In this experiment, we assume $X=[S,V]^T$, and firstly sample $S$ from mixture Gaussian distribution characterized as $\sum_{i=1}^k z_k \mathcal{N}(\mu_i,I)$ and the target $Y = \theta_s^TS + \beta S_1S_2S_3+\mathcal{N}(0,0.3)$. 
Then the unstable covariates $V$ are generated by anti-causal effect from $Y$ as 
\begin{equation}
	\small
	V = \theta_v Y + \mathcal{N}(0,\sigma(\mu_i)^2)
\end{equation}
where $\sigma(\mu_i)$ means the Gaussian noise added to $V$ depends on which component the stable covariates $S$ belong to. 
Intuitively, in different Gaussian components, the corresponding correlations between $V$ and $Y$ are varying due to the different value of $\sigma(\mu_i)$. 
The larger the $\sigma(\mu_i)$ is, the weaker correlation between $V$ and $Y$. 
We use the mixture weight $Z=[z_1,\dots,z_k]^T$ to define different environments, where different mixture weights represent different overall strength of the effect $Y$ on $V$.

In this experiment, we set $\beta=0.1$ and build 10 environments with varying $\sigma$ and the dimension of $S,V$, the first three for training and the last seven for testing. 
The average prediction errors are shown in Table \ref{tab:anti-causal}, where the first three environments are used for training and the last seven are not captured in training with weaker correlation between $V$ and $Y$.
{\bf ERM} and {\bf IRM} achieve the best training performance with respect to their prediction errors on training environments $e_1,e_2,e_3$, while their performances in testing are poor. 
{\bf WDRL} performs worst due to its over pessimism problem. {\bf SAL} achieves nearly uniformly good performance in training environments as well as the testing ones, which validates the effectiveness of our method and proves the excellent generalization ability of SAL.

\begin{table*}[htbp]
	\small
	\centering
	
	\vskip 0.05in
	
	\begin{tabular}{|l|c|c|c|c|c|c|c|c|c|c|}
		\hline
		\multicolumn{11}{|c|}{\textbf{Scenario 1: $S\in R^5,\ V\in R^5$}}\\
		\hline
		$e$&\multicolumn{3}{|c|}{Training environments}&\multicolumn{7}{|c|}{Testing environments}\\
		\hline
		Methods &  $e_1$ & $e_2$ &$e_3$ & $e_4$ &  $e_5$ & $e_6$ &$e_7$ & $e_8$  & $e_9$ & $e_{10}$  \\
		\hline 
		ERM & \bf 0.281 & 0.305 & 0.341 & 0.461 & 0.555 & 0.636&  0.703& 0.733 & 0.765 & 0.824  \\
		IRM & 0.287 &\bf 0.293 &\bf 0.329 &\bf 0.345 & 0.382 & 0.420 & 0.444 & 0.461 & 0.478 & 0.504 \\
		WDRL&0.282 & 0.331 & 0.399 & 0.599 & 0.750 & 0.875 & 0.983 & 1.030 & 1.072 & 1.165 \\
		\hline
		SAL &0.324 & 0.329 & 0.331 &  0.357 &\bf 0.380 & \bf 0.403 &\bf 0.425 &\bf 0.435 &\bf 0.446 &\bf 0.458 \\
		\hline
		
		\multicolumn{11}{|c|}{\textbf{Scenario 2: $S\in R^9,\ V\in R^1$}}\\
		\hline
		$e$&\multicolumn{3}{|c|}{Training environments}&\multicolumn{7}{|c|}{Testing environments}\\
		\hline
		Methods &  $e_1$ & $e_2$ &$e_3$ & $e_4$ &  $e_5$ & $e_6$ &$e_7$ & $e_8$  & $e_9$ & $e_{10}$  \\
		\hline 
		ERM &\bf 0.272 &\bf  0.278 & 0.298  & 0.362 & 0.411 &
  0.460 &  0.504 & 0.526&  0.534 &  0.580  \\
		IRM & 0.306 &  0.312 & 0.325  & 0.328 & 0.343 &  0.358 & 0.365 & 0.374&  0.377 & 0.394 \\
		WDRL&0.300 & 0.314 & 0.332  & 0.396 & 0.441 &
  0.483 &  0.529& 0.545 &0.555 &0.596\\
		\hline
		SAL &0.290 & 0.284 &\bf 0.288  &\bf 0.287 &\bf 0.288 &\bf
  0.287 &\bf  0.290 &\bf 0.284 &\bf 0.293&\bf  0.294\\
		\hline
	\end{tabular}
\caption{Results of the anti-causal effect experiment. The average prediction errors of 15 runs are reported.}
\label{tab:anti-causal}
\end{table*}

\subsection{Real Data}
\subsubsection{Regression}

\begin{figure*}[!ht]
\small
\vskip -0.1in
     \begin{minipage}{0.65\textwidth}
       \subfigure[$Mean\_Error$ and $Std\_Error$.]{\label{img:house-summary}\includegraphics[width=0.49\textwidth]{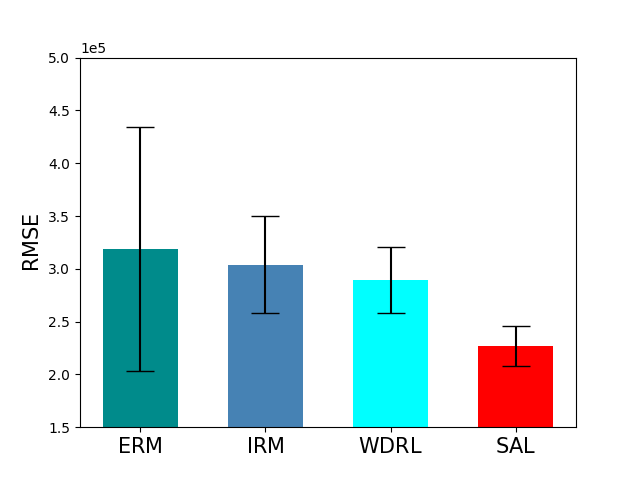}}
       \hfill
		\subfigure[Prediction error with respect to build year.]{\label{img:house-month}\includegraphics[width=0.49\textwidth]{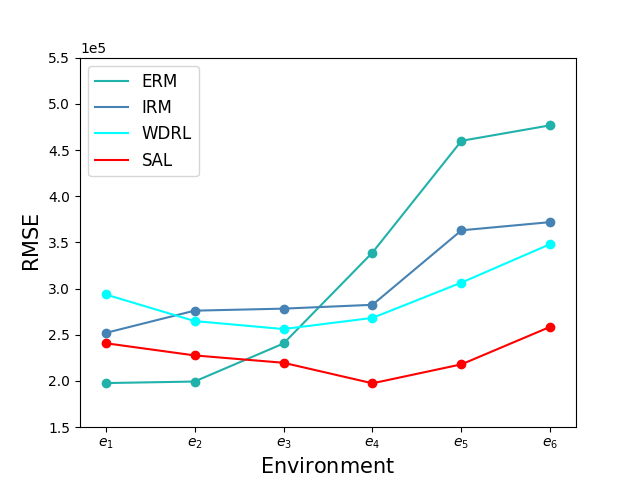}}       
		\caption{Results of the real regression dataset. $\mathrm{RMSE}$ refers to the Root Mean Square Error.}
     \end{minipage}
     \hfill
     \begin{minipage}{0.35\textwidth}
     
       \includegraphics[width=\textwidth]{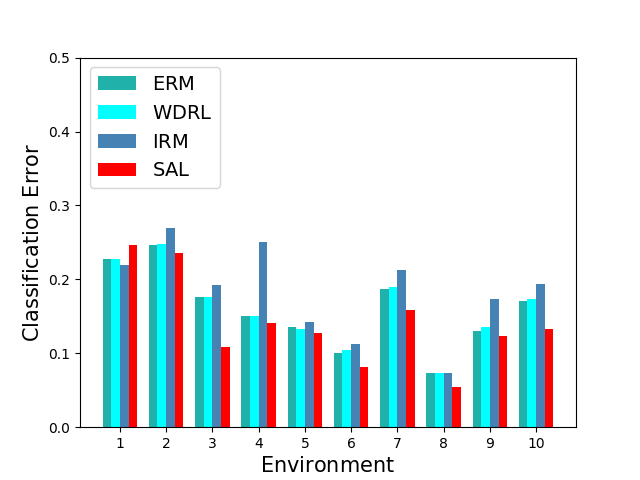}
       \caption{Results of the Adult dataset.}
       \label{img:adults}
     \end{minipage}
     \vskip -0.1in
\end{figure*}

In this experiment, we use a real-world regression dataset (Kaggle) of house sales prices from King County, USA, which includes the houses sold between May 2014 and May 2015 \footnote{https://www.kaggle.com/c/house-prices-advanced-regression-techniques/data}. 
The target variable is the transaction price of the house and each sample contains 17 predictive variables such as the built year of the house, number of bedrooms, and square footage of home, etc. 
We normalize all the predictive covariates to get rid of the influence by their original scales.

To test the stability of different algorithms, we simulate different environments according to the built year of the house.
It is fairly reasonable to assume the correlations between parts of the covariates and the target may vary along time, due to the changing popular style of architecture. 
Specifically, the houses in this dataset were built between $1900\sim 2015$ and we split the dataset into 6 periods, where each period approximately covers a time span of two decades. 
In training, we train all methods on the first and second decade where $built\ year \in [1900,1910)\ and\ [1910,1920)$ respectively and validate on 100 data points sampled from the second period. 

From the results shown in figure \ref{img:house-summary}, we can find that {\bf SAL} achieves not only the smallest $\mathrm{Mean\_Error}$ but also the lowest $\mathrm{Std\_Error}$ compared with baselines. 
From figure \ref{img:house-month}, we can find that from period 4 and so on, where large distribution shifts occurs, \textbf{ERM} performs poorly and has larger prediction errors. 
\textbf{IRM} performs stably across the first 4 environments but it also fails on the last two, whose distributional shifts are stronger. 
\textbf{WDRL} maintains stable across environments while the mean error is high, which is consistent with our analysis in \ref{exp:toy} that WDRL equally perturbs all covariates and sacrifices accuracy for robustness. 
From figure \ref{img:house-month}, we can find that from period 3 and so on, \textbf{SAL} performs better than ERM, IRM and WDRL, especially when distributional shifts are large. 
In periods 1-2 with slight distributional shift, the SAL method incurs a performance drop compared with IRM and WDRL, while SAL performs much better when larger distributional shifts occur, which is consistent with our intuition that our method sacrifice a little performance in nearly I.I.D. setting for its superior robustness under unknown distribution shifts.

\subsubsection{Classification}

Finally, we validate the effectiveness of our SAL on classification tasks, including an income prediction task and colored MNIST classification task.

{\bf Income Prediction} In this task we use the Adult dataset\cite{Dua:2019} which involves predicting personal income levels as above or below \$50,000 per year based on personal details. 
We split the dataset into 10 environments according to demographic attributes, among which distributional shifts might exist.
In training phase, we train all methods on 693 data points from environment 1 and 200 points from the second respectively and validate on 100 points sampled from both.
We normalize all the predictive covariates to get rid of the influence by their original scales. 
In testing phase, we test all methods on the 10 environments and report the mis-classification rate on all environments in figure \ref{img:adults}.
From the results shown in figure \ref{img:adults}, we can find that the \textbf{SAL} outperforms baselines on almost all environments except a slight drop on the first.
However, our SAL outperforms the others in the rest 8 environments where agnostic distributional shifts occur. 

{\bf Colored MNIST} 
In this task we build a synthetic binary classification task derived from MNIST.
The goal is to predict a binary label assigned to each image based on the digit. 
We color each image either red or green which spuriously correlates with the label similar to \cite{arjovsky2019invariant}.
The direction of the correlation is reversed in the testing environment, which ruins the method relying on such spurious correlation to predict. 
Specifically, we generate the color id by flipping the label with probability $\mu$, where $\mu=1.0$ in the first environment, $\mu=0.3$ in the second and $\mu=-1.0$ in testing.
Furthermore, we induce noisy labels by randomly flipping the label with probability 0.2. 

In this experiment, we consider the imbalanced mixture which is a more challenging and practical problem. 
Specifically, we sample 20000 images from environment 1 and 500 from 2 as training data and 10000 images from environment 3 for testing.
For our SAL and WDRL, we conduct a two-stage optimization which firstly uses a three-layer CNN to extract the representation of 128 dimensions as the input covariates. 
For ERM and IRM, we use the same architecture and do the end-to-end optimization.
We select the hyper-parameters according to the performance on the validation set sampled from training environments.
From the results in Table \ref{tab:mnist}, ERM performs terribly because of the spurious correlations and IRM and WDRL are closed to random guess.
Our SAL outperforms all baselines, which shows that our method can handle more complicated data such as vision and lingual data with a feature extractor(e.g. deep neural network).

\begin{table}[htbp]
	\centering  
	\small
	\begin{tabular}{|c|c|c|c|c|c|}  
		\hline 
		Algorithm & ERM & WDRL &IRM & SAL & Random \\  
		\hline
		Test Acc &0.085&0.48&0.51&\bf 0.57& 0.50 \\
		\hline
	\end{tabular}
	\caption{Results of the colored MNIST experiment. We report the average results of 10 runs.}  
	\label{tab:mnist}
	\vskip -0.1in
\end{table}

\section{Conclusion}
In this paper, we address a practical problem of overwhelmingly-large uncertainty set in robust learning, which often results in unsatisfactory performance under distributional shifts in real situations. 
We propose the Stable Adversarial Learning (SAL) algorithm that anisotropically considers each covariate to achieve more realistic robustness. 
We theoretically show that our method constructs a better uncertainty set. 
Empirical studies validate the effectiveness of our methods in terms of uniformly good performance across different distributed data.
We temporarily focus our method at raw feature level for solid theoretical guarantees, 
and we leave the extension of combining representation learning into our framework as the future work.

\newpage
\section{Acknowledgements}
This work was supported in part by National Key R\&D Program of China (No. 2018AAA0102004), National Natural Science Foundation of China (No. U1936219, 61772304, 61531006, U1611461), Beijing Academy of Artificial Intelligence (BAAI ), and a grant from the Institute for Guo Qiang, Tsinghua University.
Kun Kuang's research was supported in part by National Natural Science Foundation of China (No. 62006207), National Key Research and Development Program of China (No. 2018AAA0101900), the Fundamental Research Funds for the Central Universities.
Bo Li’s research was supported by the Tsinghua University Initiative Scientific Research Grant, No. 2019THZWJC11; National Natural Science Foundation of China, No. 71490723 and No. 71432004; Science Foundation of Ministry of Education of China, No. 16JJD630006.

\bibliography{aaai}

\begin{thebibliography}{28}
\providecommand{\natexlab}[1]{#1}
\providecommand{\url}[1]{\texttt{#1}}
\providecommand{\urlprefix}{URL }
\expandafter\ifx\csname urlstyle\endcsname\relax
  \providecommand{\doi}[1]{doi:\discretionary{}{}{}#1}\else
  \providecommand{\doi}{doi:\discretionary{}{}{}\begingroup
  \urlstyle{rm}\Url}\fi

\bibitem[{Arjovsky et~al.(2019)Arjovsky, Bottou, Gulrajani, and
  Lopez-Paz}]{arjovsky2019invariant}
Arjovsky, M.; Bottou, L.; Gulrajani, I.; and Lopez-Paz, D. 2019.
\newblock Invariant risk minimization.
\newblock \emph{arXiv preprint arXiv:1907.02893} .

\bibitem[{Bartlett and Mendelson(2002)}]{bartlett2002rademacher}
Bartlett, P.~L.; and Mendelson, S. 2002.
\newblock Rademacher and Gaussian complexities: Risk bounds and structural
  results.
\newblock \emph{Journal of Machine Learning Research} 3(Nov): 463--482.

\bibitem[{Ben-Tal and Nemirovski(1998)}]{ben1998robust}
Ben-Tal, A.; and Nemirovski, A. 1998.
\newblock Robust convex optimization.
\newblock \emph{Mathematics of operations research} 23(4): 769--805.

\bibitem[{Berk et~al.(2018)Berk, Heidari, Jabbari, Kearns, and
  Roth}]{berk2018fairness}
Berk, R.; Heidari, H.; Jabbari, S.; Kearns, M.; and Roth, A. 2018.
\newblock Fairness in criminal justice risk assessments: The state of the art.
\newblock \emph{Sociological Methods \& Research} 0049124118782533.

\bibitem[{Bhattad et~al.(2019)Bhattad, Chong, Liang, Li, and
  Forsyth}]{DBLP:journals/corr/abs-1904-06347}
Bhattad, A.; Chong, M.~J.; Liang, K.; Li, B.; and Forsyth, D.~A. 2019.
\newblock Big but Imperceptible Adversarial Perturbations via Semantic
  Manipulation.
\newblock \emph{CoRR} abs/1904.06347.
\newblock \urlprefix\url{http://arxiv.org/abs/1904.06347}.

\bibitem[{Daume and Marcu(2006)}]{daume2006domain}
Daume, H.; and Marcu, D. 2006.
\newblock Domain adaptation for statistical classifiers.
\newblock \emph{Journal of Artificial Intelligence Research} 26(1): 101--126.

\bibitem[{Dua and Graff(2017)}]{Dua:2019}
Dua, D.; and Graff, C. 2017.
\newblock {UCI} Machine Learning Repository.
\newblock \urlprefix\url{http://archive.ics.uci.edu/ml}.

\bibitem[{Duchi and Namkoong(2018)}]{duchi2018learning}
Duchi, J.; and Namkoong, H. 2018.
\newblock Learning models with uniform performance via distributionally robust
  optimization.
\newblock \emph{arXiv preprint arXiv:1810.08750} .

\bibitem[{Esfahani and Kuhn(2018)}]{esfahani2018data}
Esfahani, P.~M.; and Kuhn, D. 2018.
\newblock Data-driven distributionally robust optimization using the
  Wasserstein metric: Performance guarantees and tractable reformulations.
\newblock \emph{Mathematical Programming} 171(1-2): 115--166.

\bibitem[{Frogner et~al.(2019)Frogner, Claici, Chien, and
  Solomon}]{frogner2019incorporating}
Frogner, C.; Claici, S.; Chien, E.; and Solomon, J. 2019.
\newblock Incorporating Unlabeled Data into Distributionally Robust Learning.
\newblock \emph{arXiv preprint arXiv:1912.07729} .

\bibitem[{Goodfellow, Shlens, and Szegedy(2014)}]{goodfellow2014explaining}
Goodfellow, I.~J.; Shlens, J.; and Szegedy, C. 2014.
\newblock Explaining and harnessing adversarial examples.
\newblock \emph{arXiv preprint arXiv:1412.6572} .

\bibitem[{Huval et~al.(2015)Huval, Wang, Tandon, Kiske, Song, Pazhayampallil,
  Andriluka, Rajpurkar, Migimatsu, Cheng-Yue et~al.}]{huval2015empirical}
Huval, B.; Wang, T.; Tandon, S.; Kiske, J.; Song, W.; Pazhayampallil, J.;
  Andriluka, M.; Rajpurkar, P.; Migimatsu, T.; Cheng-Yue, R.; et~al. 2015.
\newblock An empirical evaluation of deep learning on highway driving.
\newblock \emph{arXiv preprint arXiv:1504.01716} .

\bibitem[{Kuang et~al.(2018)Kuang, Cui, Athey, Xiong, and Li}]{kuang2018stable}
Kuang, K.; Cui, P.; Athey, S.; Xiong, R.; and Li, B. 2018.
\newblock Stable prediction across unknown environments.
\newblock In \emph{Proceedings of the 24th ACM SIGKDD International Conference
  on Knowledge Discovery \& Data Mining}, 1617--1626.

\bibitem[{Kuang et~al.(2020)Kuang, Xiong, Cui, Athey, and Li}]{kuang2020stable}
Kuang, K.; Xiong, R.; Cui, P.; Athey, S.; and Li, B. 2020.
\newblock Stable prediction with model misspecification and agnostic
  distribution shift.
\newblock In \emph{Proceedings of the AAAI Conference on Artificial
  Intelligence}, volume~34, 4485--4492.

\bibitem[{Kukar(2003)}]{kukar2003transductive}
Kukar, M. 2003.
\newblock Transductive reliability estimation for medical diagnosis.
\newblock \emph{Artificial Intelligence in Medicine} 29(1-2): 81--106.

\bibitem[{Madry et~al.(2017)Madry, Makelov, Schmidt, Tsipras, and
  Vladu}]{madry2017towards}
Madry, A.; Makelov, A.; Schmidt, L.; Tsipras, D.; and Vladu, A. 2017.
\newblock Towards deep learning models resistant to adversarial attacks.
\newblock \emph{arXiv preprint arXiv:1706.06083} .

\bibitem[{Papernot et~al.(2016)Papernot, McDaniel, Jha, Fredrikson, Celik, and
  Swami}]{papernot2016limitations}
Papernot, N.; McDaniel, P.; Jha, S.; Fredrikson, M.; Celik, Z.~B.; and Swami,
  A. 2016.
\newblock The limitations of deep learning in adversarial settings.
\newblock In \emph{2016 IEEE European symposium on security and privacy
  (EuroS\&P)}, 372--387. IEEE.

\bibitem[{Rudin and Ustun(2018)}]{rudin2018optimized}
Rudin, C.; and Ustun, B. 2018.
\newblock Optimized scoring systems: Toward trust in machine learning for
  healthcare and criminal justice.
\newblock \emph{Interfaces} 48(5): 449--466.

\bibitem[{Sagawa et~al.(2019)Sagawa, Koh, Hashimoto, and
  Liang}]{sagawa2019distributionally}
Sagawa, S.; Koh, P.~W.; Hashimoto, T.~B.; and Liang, P. 2019.
\newblock Distributionally Robust Neural Networks for Group Shifts: On the
  Importance of Regularization for Worst-Case Generalization.
\newblock \emph{arXiv preprint arXiv:1911.08731} .

\bibitem[{Sagawa et~al.(2020)Sagawa, Raghunathan, Koh, and Liang}]{2020An}
Sagawa, S.; Raghunathan, A.; Koh, P.~W.; and Liang, P. 2020.
\newblock An Investigation of Why Overparameterization Exacerbates Spurious
  Correlations .

\bibitem[{Shen et~al.(2018)Shen, Cui, Kuang, Li, and Chen}]{2018Causally}
Shen, Z.; Cui, P.; Kuang, K.; Li, B.; and Chen, P. 2018.
\newblock Causally Regularized Learning with Agnostic Data Selection Bias.
\newblock In \emph{2018 ACM Multimedia Conference}.

\bibitem[{Shen et~al.(2020)Shen, Cui, Liu, Zhang, Li, and
  Chen}]{shen2020stable}
Shen, Z.; Cui, P.; Liu, J.; Zhang, T.; Li, B.; and Chen, Z. 2020.
\newblock Stable learning via differentiated variable decorrelation.
\newblock In \emph{Proceedings of the 26th ACM SIGKDD International Conference
  on Knowledge Discovery \& Data Mining}, 2185--2193.

\bibitem[{Shen et~al.(2019)Shen, Cui, Zhang, and Kuang}]{shen2019stable}
Shen, Z.; Cui, P.; Zhang, T.; and Kuang, K. 2019.
\newblock Stable Learning via Sample Reweighting.
\newblock \emph{arXiv: Learning} .

\bibitem[{Sinha, Namkoong, and Duchi(2018)}]{SinhaCertifying}
Sinha, A.; Namkoong, H.; and Duchi, J. 2018.
\newblock Certifying Some Distributional Robustness with Principled Adversarial
  Training.
\newblock \emph{International Conference on Learning Representations} .

\bibitem[{Torralba and Efros(2011)}]{torralba2011unbiased}
Torralba, A.; and Efros, A.~A. 2011.
\newblock Unbiased look at dataset bias 1521--1528.

\bibitem[{Vaishnavi et~al.(2019)Vaishnavi, Cong, Eykholt, Prakash, and
  Rahmati}]{vaishnavi2019can}
Vaishnavi, P.; Cong, T.; Eykholt, K.; Prakash, A.; and Rahmati, A. 2019.
\newblock Can Attention Masks Improve Adversarial Robustness?
\newblock \emph{arXiv preprint arXiv:1911.11946} .

\bibitem[{Wah et~al.(2011)Wah, Branson, Welinder, Perona, and
  Belongie}]{wah2011caltech}
Wah, C.; Branson, S.; Welinder, P.; Perona, P.; and Belongie, S. 2011.
\newblock The caltech-ucsd birds-200-2011 dataset .

\bibitem[{Ye and Zhu(2018)}]{ye2018bayesian}
Ye, N.; and Zhu, Z. 2018.
\newblock Bayesian adversarial learning.
\newblock In \emph{Proceedings of the 32nd International Conference on Neural
  Information Processing Systems}, 6892--6901. Curran Associates Inc.

\end{thebibliography}
\end{document}